\pdfoutput=1

\documentclass[11pt]{article}

\usepackage[]{EMNLP2023}

\usepackage{times}
\usepackage{latexsym}

\usepackage{graphicx}
\usepackage{subcaption}
\usepackage{amsmath}
\usepackage{CJKutf8}

\usepackage[T1]{fontenc}

\usepackage[utf8]{inputenc}

\usepackage{microtype}

\usepackage{inconsolata}

\title{WMT24 Test Suite: Gender Resolution in Speaker-Listener Dialogue Roles}

\author{Hillary Dawkins \And Isar Nejadgholi  \\
  Digital Technologies Research Centre\\National Research Council Canada (NRC-CNRC) \\
  \texttt{\{hillary.dawkins, isar.nejadgholi, chikiu.lo\}@nrc-cnrc.gc.ca} \\
  \And Chi-kiu Lo \begin{CJK*}{UTF8}{bsmi}羅致翹\end{CJK*}}

\begin{document}
\maketitle
\begin{abstract}
We assess the difficulty of gender resolution in literary-style dialogue settings and the influence of gender stereotypes. Instances of the test suite contain spoken dialogue interleaved with external meta-context about the characters and the manner of speaking. We find that character and manner stereotypes outside of the dialogue significantly impact the gender agreement of referents within the dialogue. \href{https://github.com/hillary-dawkins/wmt24-gender-dialogue}{https://github.com/hillary-dawkins/wmt24-gender-dialogue}.    
\end{abstract}

\section{Introduction}

Gender bias and gender effects in machine translation are prevalent in translation directions where gender relevancy increases from source to target language \cite{savoldi-etal-2021-gender, barclay2024investigatingmarkersdriversgender, savoldi-etal-2023-test}. 
English has minimal morphological effects caused by natural gender, whereas many languages (e.g. French, Spanish, Czech, Icelandic, German) have grammatical gender cases for various parts of speech which sometimes need to align with natural gender for animate nouns. 
For example ``I am happy'' in the source language English has divergent translations in the target language French (``Je suis heureux/heureuse'') depending on the natural gender of the speaker.   
The consequence is that gender-alignment errors can easily arise in such translation directions. 
Furthermore, stereotypes are known to drive gender agreement (e.g., systems may tend to prefer the translation ``Je suis jolie'' over ``Je suis joli'' for ``I am pretty'' despite incomplete gender context) \cite{solmundsdottir-etal-2022-mean}, and these stereotype effects can persist even when unambiguous gender information is provided \cite{stanovsky-etal-2019-evaluating, troles-schmid-2021-extending, kocmi-etal-2020-gender}.  

Typically, these gender effects are studied in isolation or semantically-bleached settings (as in the above examples). There it is known that the internal characteristics of adjective words, such as the gender stereotype, sentiment, and type (appearance or character), are significant factors influencing the choice of gender agreement in translation \cite{solmundsdottir-etal-2022-mean}.
However, the need for gender agreement also occurs in more complex settings, such as over long ranges, and passages involving multiple potential referents.    

Due to increasing interest in paragraph-level translation and literary domains, here we assess the challenge of speaker-listener role resolution in literary dialogue settings. In particular, the gender of the speaker and listener must be resolved correctly to obtain a correct translation, and we suppose that gender stereotype effects can further add to the task difficulty. We find that stereotypical character descriptions and manners of speaking are significant influences on the gender alignment, generally overshadowing the internal adjective traits.  

\section{Test Suite Description}
\label{sec:test_suite}

This test suite measures the gender resolution tendencies of machine translation systems in literary-style dialogue settings. 
In this setting, spoken dialogue (in quotations or otherwise delimited) is interleaved with meta-context about the dialogue (e.g., the speaker, the listener(s), and character and environment descriptions). 
When spoken dialogue refers to a person, a challenge arises in resolving the referent given the meta-context. 
The test suite includes three target languages (Spanish, Czech, and Icelandic), where the gender of the referent affects the correct translation.

Here, we focus on two-person conversations, where adjectives are used within dialogue to describe either the speaker or the listener. Within a single source passage, both characters may take on both the speaker and listener roles at times. Since adjectives are gender-neutral in the source language (English), the gender of the adjective's referent must be determined from the meta-context, if possible. 
The test suite contains inputs where the gender remains unknown given the complete context (termed gender-ambiguous cases), and inputs where the gender can be unambiguously resolved given the complete context (termed gender-determined cases). 

The test suite contains a handful of template types (each detailed in Appendix \ref{sec:appendix_templates}) to assess the influence of stereotype cues in the meta-context and the structural features of the passage. Stereotype features include character descriptions and the manner of speaking (controlled using adverbs). Structural features include the number of referents in a single passage, partial or complete gender information, first- or third-person speakers, and adjective repetition. Some challenging features of the templates include adjectives that appear before the referent is introduced, and repeated adjectives referring to different entities. 

The templates use vocabulary sets for adjectives ($n=350$), gender-stereotyped adverbs ($n=29$), and gender-stereotyped occupation words ($n=44$). 
Each adjective is labeled with its gender stereotype (M/F/neutral), sentiment (positive/negative/neutral), and type (character/appearance).  
The full vocabulary set with annotations is released as part of the test suite contribution. 

\section{Methodology}
\label{sec:methods}

The adjective translations are extracted from the target languages and processed using dictionary searches\footnote{https://bin.arnastofnun.is/ \\ https://islenskordabok.arnastofnun.is/ \\ https://slovnik.seznam.cz/ \\ https://dictionaryapi.com/products/api-spanish-dictionary \\ https://en.wiktionary.org/ \\ https://cs.wiktionary.org/} to obtain the gender agreement label. 
The advantage of using dictionary searches over automated morphological gender taggers is that irregular adjectives (e.g. ``rosa'' in Spanish) are correctly classified, and the use of different parts of speech or out-of-dictionary words can also be monitored. For example, the use of a gender-neutral noun phrase or direct substitution of an English word should be counted as a neutral label for our purposes. Only when a translated word is not found in any dictionary search, is it passed to auto gender-tagging based on its morphological features (e.g. an ``o'' vs. an ``a'' ending in Spanish). This second pass allows for (possibly hallucinated) out-of-dictionary words to be included in the analysis, but only if they strongly resemble a regular adjective form (e.g. ``víktur'' in Icelandic may be derived from the English source word ``victorious'', but clearly a masculine adjective ending has been chosen in translation). A small portion of words remains unclassified after both passes are complete, meaning that they neither exist in the dictionary nor resemble a regular adjective in the target language. The fine-grained annotations for each extracted translation, in addition to the final gender label (one of M, F, N, or unclassified), are released with the test suite results for further analysis.   

The scope of analysis in this paper is limited to the subset of M- and F-labeled translations. That is, when a gendered adjective form \textit{is} chosen by a translation system, we are interested in the factors that influence this choice, and the corresponding translation errors that occur when an adjective form does not match the referent's gender.
To this end, results throughout the paper are presented in three ways. 

When the gender of a referent is unknown, we report the proportion of masculine and feminine adjective declensions to observe the system's tendencies in ambiguous settings. When the gender of a referent is known, we report the accuracy of the adjective declensions. Typically, the underlying effect (e.g., the influence of stereotypes) is the same in both cases. However, it is important to know that the effect persists even when unambiguous gender context is available. Both proportion and accuracy results are always reported using balanced subsets\footnote{Adjective traits are balanced on type and sentiment, and exclude gender stereotypes; structural factors such as listener and speaker roles are balanced as applicable to the template type.} of the relevant test suite subset.    

Lastly, we wish to understand the relative importance of factors that influence the system's choice of gender agreement. 
To do so, we perform regression analyses where the dependent variable to predict is the gender declension of the translation, and independent variables include both internal adjective factors (the gender stereotype, sentiment, and type), and external factors that are introduced through the meta context (e.g. character descriptions). The regression coefficients are reported with significance levels. 

\section{Gender-Stereotyped Manner}
\label{sec:results_manner}

\begin{figure}
  \centering
  \includegraphics[scale = .234, clip]{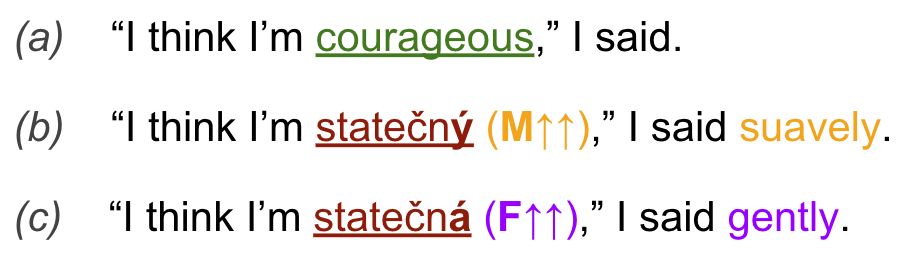}
  \caption{Gender-stereotyped adverbs outside of the dialogue affect the adjective's gender agreement with the speaker within the dialogue. Source sentences in English include instances without adverbs (a) and with stereotypically masculine (b) or feminine adverbs (c). When translated to the target language, adjectives tend to align with the stereotype (adjectives shown here in Czech).}
\label{fig:adverb}
\end{figure}

Firstly, we observe that the manner of speaking in literary dialogue settings can significantly affect the gender prediction of the speaker. Furthermore, this influence is susceptible to gender stereotypes. Refer to the example shown in Figure \ref{fig:adverb}. 

Within the \textbf{Stereo-Adverb} test suite subset, all adjectives refer to a first-person speaker (I), and therefore the natural gender of the adjective's referent is ambiguous in the source language. We report the proportion of male declensions on subsets (a) with no adverb, (b) a male-stereotyped adverb, and (c) a female-stereotyped adverb (full results in Appendix \ref{sec:appendix_all_results}). The majority of systems display a difference greater than 10\% when the adverb switches from male- to female-stereotyped. The systems with the largest effects are shown in Table \ref{table:top_adverbs}. Note that the most affected systems include those that defy the usual default-male agreement in ambiguous gender cases in the baseline setting (i.e., in the absence of any adverb). Here we see that the default-female agreement is unstable with respect to stereotype cues.   

To compare the influence of speaking manner to the influence of internal adjective traits, we perform regression analysis where the dependent variable to predict is the gender declension. Independent variables are the gender stereotype label of the adverb, and the gender stereotype, the sentiment, and the type (appearance or character) of the adjective. The analysis shows that adverb influence is comparable or stronger than these internal adjective characteristics within this test suite (see Table \ref{tab:adverb_reg_es}). 

\begin{table*}
\centering
\begin{tabular}{l c c c c c c c}
\hline
\textbf{System} & \textbf{$F$} & \textbf{$M$} & \textbf{$F_{M}$} & \textbf{$M_{M}$} & \textbf{$F_{F}$} & \textbf{$M_{F}$} & \textbf{$\Delta M_{M-F}$} \\
\hline
CUNI-MH	&	0.703	&	0.297	&	0.379	&	0.621	&	0.950	&	0.050	&	\textbf{0.571}	\\
ONLINE-W	&	0.591	&	0.409	&	0.387	&	0.613	&	0.884	&	0.116	&	\textbf{0.497}	\\
CommandR-plus	&	0.340	&	0.660	&	0.120	&	0.880	&	0.554	&	0.446	&	\textbf{0.434}	\\
Aya23	&	0.370	&	0.631	&	0.187500	&	0.813	&	0.612	&	0.388	&	\textbf{0.425}	\\
\hline
\end{tabular}
    \caption{\textbf{Gender-Stereotyped Manner}: The proportion of adjectives with male ($M$) and female ($F$) agreement on the \textbf{Stereo-Adverb} test suite subset for the most affected translation systems in the English to Czech translation direction. All adjectives self-refer to the speaker of unknown gender. Subscripts ($M$ and $F$) denote the use of gender-stereotyped adverbs to control the manner of speaking (e.g., $M_F$ denotes the proportion of adjectives with a male declension within instances using a stereotypically feminine adverb, as shown in Figure \ref{fig:adverb} example (c)). The unsubscripted results refer to no adverb (as shown in Figure \ref{fig:adverb} example (a)). The overall strength of the adverb effect is reported using the difference $\Delta M_{M-F}$.}
    \label{table:top_adverbs}
\end{table*}

\section{Gender-Stereotyped Characters}
\label{sec:results_char}

\begin{figure*}
  \centering
  \begin{subfigure}{\textwidth}
         \centering
         \includegraphics[scale = .25, clip]{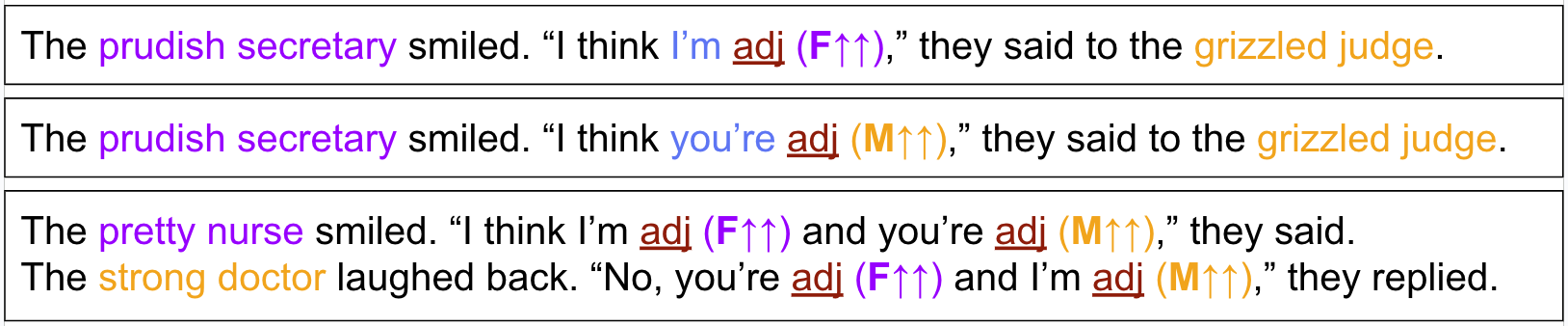}
         \caption{Ambiguous cases: Adjectives refer to characters of unknown gender.}
         \label{fig:stereo_char_amb}
     \end{subfigure}
     \par\bigskip
  \begin{subfigure}{\textwidth}
         \centering
         \includegraphics[scale = .25, clip]{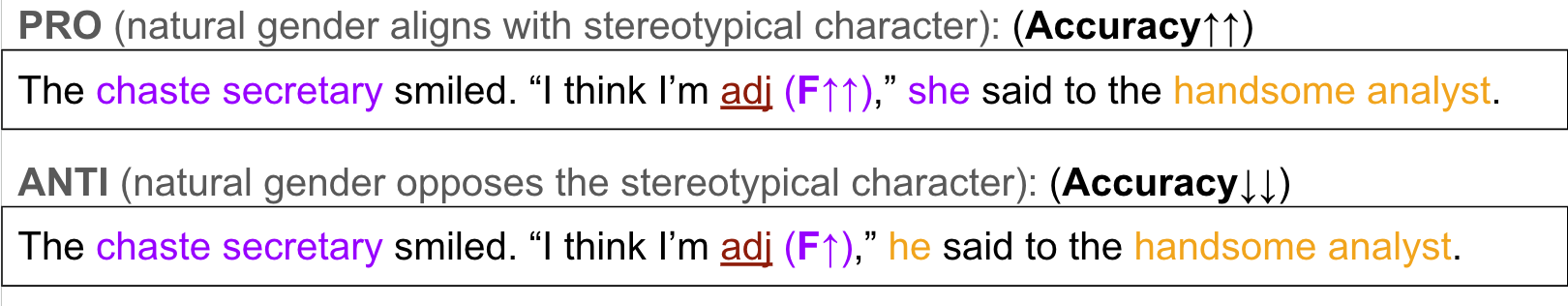}
         \caption{Determined cases: Adjectives refer to characters of known gender. The known gender either aligns with the stereotype (PRO) or opposes the stereotype (ANTI).}
         \label{fig:stereo_char_det}
     \end{subfigure}
  \caption{Gender-stereotyped character descriptions outside of the dialogue affect the adjective's gender agreement.}
\label{fig:stereo_characters}
\end{figure*}

Secondly, we observe that character descriptions that align with socially held stereotypes impact gender resolution within spoken dialogue. Refer to the examples shown in Figure \ref{fig:stereo_characters}.

Within the \textbf{Stereo-Character} test suite subset, all adjectives refer to one of two characters that have been given some stereotypical descriptions using both occupations and attributive adjectives. Template variations include single-speaker dialogue, where adjectives refer to either the speaker (I) or listener (you) (see template \ref{eqn:char_single}), and two-speaker conversations where both participants are referenced by each speaker (see template \ref{eqn:char_both}). 

In ambiguous gender cases (Figure \ref{fig:stereo_char_amb}), we report the stereotype effect again by looking at the tendency of the system to choose either a female or male adjective declension depending on the referent stereotype (full results in Appendix \ref{sec:appendix_all_results}). Characters that are described by male-leaning gender stereotypes are very likely to receive a masculine adjective, whereas the use of feminine adjectives increases for female-stereotyped characters (pushing against the default-male baseline), as shown in Table \ref{tab:top_char_amb} for the most affected systems.     

\begin{table*}
\centering
\begin{tabular}{l c c c c c c c}
\hline
\textbf{System} & \textbf{$F_{M}$} & \textbf{$M_{M}$} & \textbf{$F_{F}$} & \textbf{$M_{F}$} & \textbf{$\Delta M_{M-F}$} \\
\hline
Claude-3.5	&	0.000	&	1.000	&	0.391	&	0.609	&	\textbf{0.391}	\\
CommandR-plus	&	0.012	&	0.988	&	0.401	&	0.598	&	\textbf{0.390}	\\
Aya23	&	0.122	&	0.878	&	0.429	&	0.571	&	\textbf{0.307}	\\
Unbabel-Tower70B	&	0.058	&	0.942	&	0.359	&	0.640846	&	\textbf{0.302}	\\
GPT-4	&	0.000	&	1.000	&	0.274	&	0.726	&	\textbf{0.274}	\\
\hline
\end{tabular}
    \caption{\textbf{Gender-Stereotyped Characters}: The proportion of adjectives with male ($M$) and female ($F$) agreement on the \textbf{Stereo-Character-Amb} test suite subset (Figure \ref{fig:stereo_char_amb}) for the most affected translation systems in the English to Spanish translation direction, partitioned by the referent's gender stereotype (denoted by subscripts). The true gender of the referent is unknown, but the choice of declension is affected by the stereotypical character description. The overall strength of the character description effect is reported by the difference $\Delta M_{M-F}$.}
    \label{tab:top_char_amb}
\end{table*}

Furthermore, we find that this effect persists in determined gender cases (Figure \ref{fig:stereo_char_det}) such that agreement accuracy can drop significantly when the actual gender opposes a socially-held stereotype. We report this observation as the difference in accuracy between the PRO and ANTI template subsets (full results in Appendix \ref{sec:appendix_all_results}). Approximately half of the tested systems are not robust to stereotype cues even when the correct, unambiguous gender context is provided. The most affected systems are shown in Table \ref{tab:top_det_char}.  
As with the stereotyped adverb effect, we perform a regression analysis to probe the relative influence of stereotypical character descriptions compared to the internal adjective factors. We find that the character descriptions are much more impactful on the adjective's gender form than the internal adjective traits within this dialogue setting (see Table \ref{tab:char_reg_es}).   

\begin{table*}
\centering
\begin{tabular}{l c c c}
\hline
\textbf{System} & \textbf{Accuracy (PRO)} & \textbf{Accuracy (ANTI)} & \textbf{$\Delta($PRO, ANTI$)$} \\
\hline
ONLINE-W	&	0.985	&	0.414	&	\textbf{0.571} \\
GPT-4	&	0.990	&	0.527	&	\textbf{0.463} \\
Aya23	&	1.000	&	0.655	&	\textbf{0.345} \\
IKUN	&	0.975	&	0.702	&	\textbf{0.273} \\
\hline
\end{tabular}
    \caption{\textbf{Gender-Stereotyped Characters}: The accuracy in gender-adjective agreement on the \textbf{Stereo-Character-Det} test suite subset (Figure \ref{fig:stereo_char_det}) for the most affected translation systems in the English to Spanish translation direction. The true gender of the character either aligns with (PRO) or opposes (ANTI) the stereotypical description. The presence of stereotypical character descriptions can significantly decrease the gender translation accuracy.}
    \label{tab:top_det_char}
\end{table*}

\section{``Opposite'' or ``Same'' Binary Gender Speaker Bias}
\label{sec:results_speaker_parity}

\begin{figure*}
  \centering
  \begin{subfigure}{\textwidth}
         \centering
         \includegraphics[scale = .25, clip]{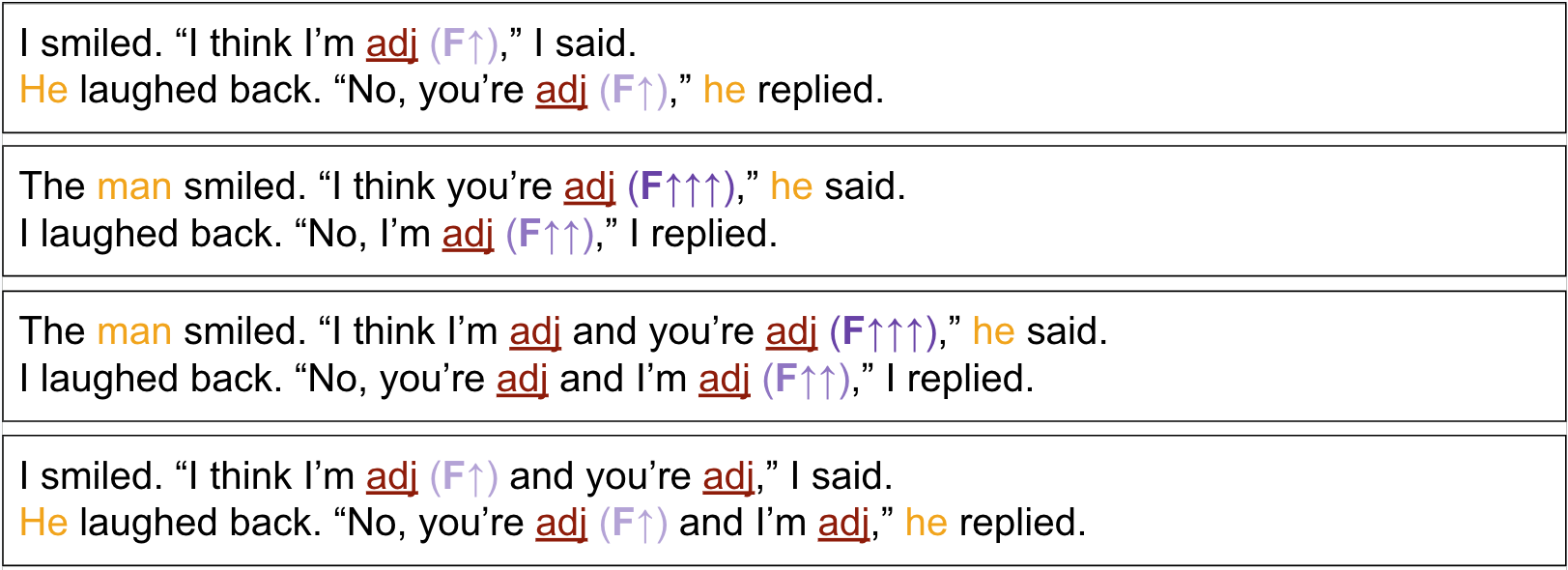}
         \caption{Ambiguous cases: Adjectives refer to a character of unknown gender, while the gender of the second character in the conversation is known (male in these examples). Adjectives referring to the gender-ambiguous character are more likely to agree with the opposite gender of the speaker (i.e., take feminine forms in these examples).}
         \label{fig:opp_amb}
     \end{subfigure}
     \par\bigskip
  \begin{subfigure}{\textwidth}
         \centering
         \includegraphics[scale = .25, clip]{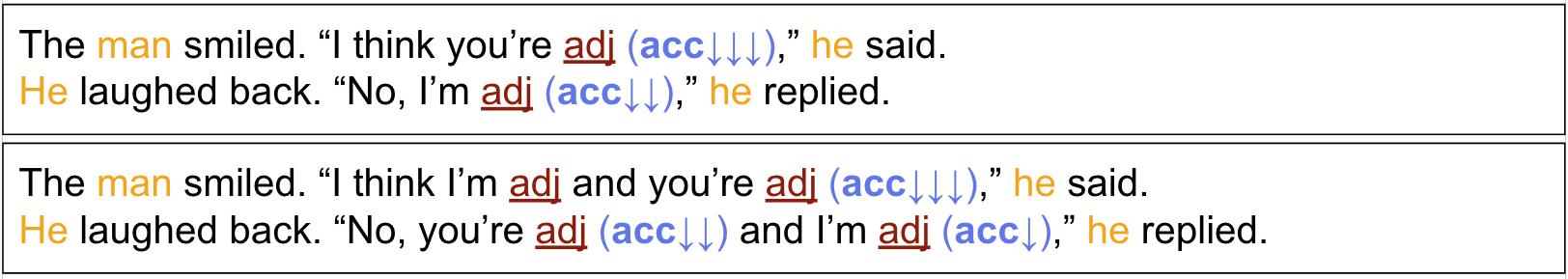}
         \caption{Determined cases where the gender of both speakers is known. Accuracy decreases for same-gender pairs due to the opposite binary gender effect.}
         \label{fig:opp_det}
     \end{subfigure}
     \par\bigskip
  \begin{subfigure}{\textwidth}
         \centering
         \includegraphics[scale = .25, clip]{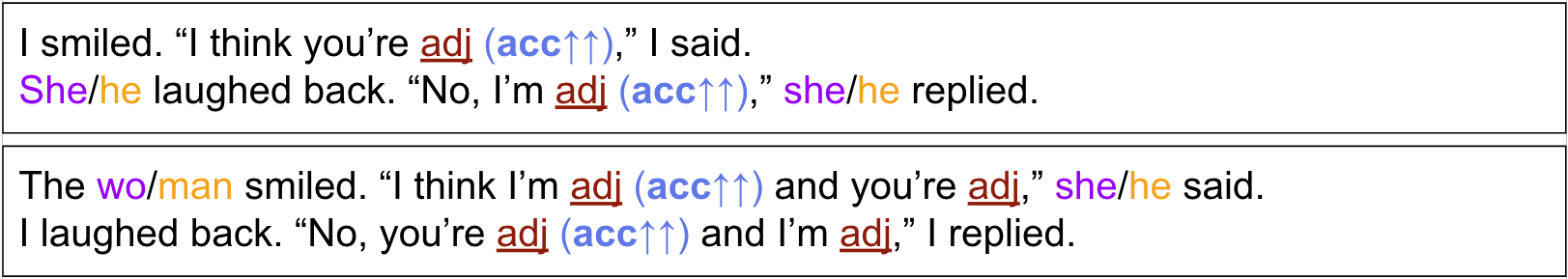}
         \caption{Determined cases where the gender of one speaker is known. Accuracy is generally high in the absence of a second gender (i.e., the opposite binary gender effect is not possible).}
         \label{fig:opp_easy}
     \end{subfigure}
  \caption{The opposite binary gender effect is present in both ambiguous (a) and determined (b) cases. Determined cases with a single known gender (c) are unchallenging despite having the same structural components (i.e. both speaker (I) and listener (you) resolutions, and need to ``look ahead'' in the text to find the adjective's referent). All effects are the same but flipped for systems that prefer same-gender speaker pairs.}
\label{fig:opp_gender}
\end{figure*}

Finally, in the absence of any gender stereotype effects, we assess the overall ``vanilla'' difficulty of this gender resolution task in non-challenge settings and the influence of different structural elements in the source input. In doing so, we observe that an ``opposite'' or same binary gender bias exists. That is, in dialogue settings between two speakers, some systems strongly predict one speaker to be male and the other female, while other systems strongly prefer same-gender pairs. This observation holds in both ambiguous and determined cases. Refer to the examples shown in Figure \ref{fig:opp_gender}. 

In ambiguous gender cases, we can observe this effect as the proportion of adjective declension choices conditioned on the known gender of the second character in the conversation (Figure \ref{fig:opp_amb}). Note that adjectives may still either refer to the speaker or listener, and both types are affected by the presence of a second known gender. Full results are shown in Appendix \ref{sec:appendix_all_results}, and a summary of the most affected systems is shown in Table \ref{tab:top_opp_amb}.  

\begin{table*}
\centering
\begin{tabular}{l c c c c c c c}
\hline
\textbf{System} & \textbf{$F_{M}$} & \textbf{$M_{M}$} & \textbf{$F_{F}$} & \textbf{$M_{F}$} & \textbf{$\Delta M_{M-F}$} \\
\hline
Claude-3.5	&	0.419	&	0.581	&	0.074	&	0.926	&	\textbf{-0.346}	\\
CommandR-plus	&	0.764	&	0.236	&	0.426	&	0.574	&	\textbf{-0.338}	\\
IKUN-C	&	0.292	&	0.708	&	0.703	&	0.297	&	\textbf{0.410}	\\
IKUN	&	0.256	&	0.744	&	0.726	&	0.274	&	\textbf{0.470}	\\
\hline
\end{tabular}
    \caption{\textbf{Opposite or Same Binary Gender Effect}: The proportion of adjectives with male ($M$) and female ($F$) agreement on the \textbf{Structure-Amb} test suite subset (Figure \ref{fig:opp_amb}) for the most affected systems in the English to Spanish translation direction. All adjectives refer to someone of an unknown gender in conversation with someone of a known gender (where that known gender is denoted by the subscripts). 
    Systems Claude-3.5 and CommandR-plus show the greatest tendency to assume opposite-gender speaker pairs ($\Delta M_{M-F} \ll 0$), and systems IOL-Research and IKUN show the greatest tendency to assume same-gender speaker pairs ($\Delta M_{M-F} \gg 0$).}  
    \label{tab:top_opp_amb}
\end{table*}

In determined gender cases, the tendency to assume either the same or opposite binary gender pairs manifests as decreased accuracy in cases that oppose this assumption. We report the accuracy in adjective agreement on test subsets where (a) only one gender is specified (Figure \ref{fig:opp_easy}), (b) both genders are specified and are opposite, (c) both genders are specified and are the same (Figure \ref{fig:opp_det}). Subset (a) is usually easiest for most systems because the same or opposite gender effect is not possible. The difference in accuracy between subsets (b) and (c) indicates the strength and direction of this effect. Full results are shown in Appendix \ref{sec:appendix_all_results} and a summary is shown in Table \ref{tab:top_opp_det}.  

\begin{table*}
\centering
\begin{tabular}{l c c c c}
\hline
\textbf{System} & \textbf{Acc (one gender)} & \textbf{Acc (same genders)} & \textbf{Acc (opp genders)} & \textbf{$\Delta($same, opp$)$} \\
\hline
CommandR-plus	&	0.987	&	0.797	&	0.991	&	\textbf{-0.194}	\\
Llama3-70B	&	0.957	&	0.806	&	0.977	&	\textbf{-0.171}	\\
ONLINE-A	&	0.734	&	0.828	&	0.668	&	\textbf{0.160}	\\
ONLINE-G	&	0.726	&	0.827	&	0.625	&	\textbf{0.202}	\\
\hline
\end{tabular}
    \caption{\textbf{Opposite or Same Binary Gender Effect}: The accuracy in gender-adjective agreement on the \textbf{Structure-Det} test suite subset (Figures \ref{fig:opp_easy} and \ref{fig:opp_det}) for the most affected systems in the English to Spanish translation direction. The second speaker in the conversation is either unknown (one gender subset), the same, or opposite to the adjective referent of known gender. Systems with an opposite binary gender effect suffer on the same-gender subset such that the difference in accuracy $\Delta($same, opp$) \ll 0$, and systems with a same-gender preference suffer on the opposite-gender subset such that the difference in accuracy $\Delta($same, opp$) \gg 0$.}
    \label{tab:top_opp_det}
\end{table*}

We note that the observed decrease in accuracy on gender pairings that oppose the system's presupposition is being driven by two features within our templates: 1. Adjectives that occur before their referent (if reading left to right), and 2. A consistency effect. For example, refer to the two examples shown in Figure \ref{fig:opp_det}. In both cases, the first adjective to translate occurs before it's referent, but after the gender of the speaker is known. Adjectives in this position are very likely to align with the same or opposite gender of the speaker in affected systems, depending on the effect direction. Following the incorrect translation of the first adjective, we observe that the adjective in the last position is likely to also be incorrect, possibly owing to a consistency effect since these refer to the same entity. 

Using regression analysis, we predict the adjective declension conditioned on structural factors: the gender of the other speaker, whether the referent is the speaker (I) or the listener (you), the gender choice in preceding adjectives that refer to the same entity (consistency), and whether the adjective occurs before the referent is introduced (``look-ahead'' position), as well as the internal traits of the adjective as always, and the true gender label for determined cases. Controlling for internal traits, the correct gender label, and the default masculine baseline, we observe that both ``look-ahead'' and referent role (listener) are influential structural factors affecting the task difficulty (refer to table \ref{tab:tab:struct_reg_es}). 


\section{Future Work}
\label{sec:future}

Here, the scope of analysis is limited to the cases where a translation system has chosen either a masculine or feminine adjective form, and ignores those cases where a neutral translation strategy was used instead. However, the labeling methodology as described in Section \ref{sec:methods} does produce a test suite with annotated neutral labels as well. The observed neutral strategies vary by target language and include the use of adjectives with the same form for the female and male gender cases (e.g. regular adjectives ending in ``e'' in Spanish, or ``í'' in Czech), the use of the neuter gender case if it exists (as in Czech and Icelandic), direct substitution of the gender-neutral source (English) adjective, the use of alternative forms (e.g. translated adjectives ending ``o/a'' in Spanish or ``(ur)'' in Icelandic), and the use of noun phrases in place of adjectives, which may be gender-neutral depending on the target language. Some of these strategies may be considered to be more correct than others (i.e. applying the neuter gender case to a person is not grammatically correct, but may still be preferred to misgendering in ambiguous cases). 

Further analysis is needed to understand how often neutral strategies are used in both the ambiguous and determined gender cases, and what factors influence a translation system's choice or ability to use a neutral strategy \cite{savoldi-etal-2024-prompt, piergentili-etal-2023-hi, lauscher-etal-2023-em}.

\section{Conclusion}

In conclusion, this test suite provides an opportunity to study the challenging task of referent resolution within literary-style dialogue settings. When spoken dialogue refers to characters described outside of dialogue in the meta-context, it adds an extra layer of complexity to the gender agreement task. 
Here we focus on language directions that are prone to gender agreement errors due to greater gender relevancy in the target language than the source language. We find that stereotypical character descriptions and manners of speaking are significant influences for some translation systems. Furthermore, some systems strongly prefer to resolve two-person conversations as same- or opposite-gender pairs. All observed effects are present in both ambiguous and determined gender cases.   

\section*{Limitations}

This test suite uses simple templates to study the influence of structural factors in a controlled manner. Although templates are varied and contain quite a few structure variables, they do not represent the diversity or complexity of real literary settings. Having identified the stereotype effects and challenge features within this test suite, future work could compile a real in-the-wild literary dialogue test suite by seeking out instances with these features of interest.   

The primary limitation of this work is the focus on binary gender. All determined gender cases within the test suite are either male or female, and the analysis of chosen gender declensions is limited to masculine and feminine forms. This is partially due to the availability of known stereotypes for binary gender, and partially due to the binary nature of gender morphology in the target languages (even if neuter grammatical gender exists, it does not apply to animate nouns). Future work should investigate the use of neutral strategies when gender is unknown as a way to avoid misgendering non-binary referents.   

\section*{Ethics Statement}

As discussed in the Limitations section, the focus on binary gender throughout the paper is a serious ethical concern, and we stress here that similar research questions are applicable to non-binary genders. We hope that the analysis presented here and the test suite results might encourage the inclusion of non-binary natural gender in future work.

\bibliography{custom}
\bibliographystyle{acl_natbib}

\appendix


\section{Test Suite Templates}
\label{sec:appendix_templates}

\subsection{Stereo-Adverb Templates}
Examples in the Stereo-Adverb test suite subset take the form:
\begin{equation}
    \text{``I think I'm } A\text{,'' I said } adverb.
\end{equation}
where $A$ ($n=130$) denotes an adjective sampled from the full adjective set, and $adverb$ can be none, $M$-stereotyped ($n=3$) or $F$-stereotyped ($n=3$). In total, there are $N = 910$ source sentences in this subset $\bigl(N = 130 \times (1+3+3) \bigr)$.

\subsection{Stereo-Character Templates}
All examples in the Stereo-Character test suite subset contain two characters that are introduced using gender-stereotyped descriptions. 
For simplicity, all character descriptions are in the form:
\begin{equation}
    C_g = a_g occ_g
\end{equation}
where $a_g$ is gender-stereotyped adjective, and $occ_g$ is a matching gender-stereotyped occupation (e.g. ``pretty nurse'' or ``strong doctor''). In each example, there is one female-stereotyped character ($n=22$) and one male-stereotyped character ($n=22$). We denote the character pairs as $(C_g, C_{\bar{g}})$. 

Templates in this test suite subset come in both single-speaker and two-way conversation styles. 
In the single-speaker template, examples are of the form:
\begin{equation}
\label{eqn:char_single}
\begin{split}
    & \text{The } C_g \text{ smiled. ``I think \{I'm, you're\} } A\text{,'' } \\
    & \text{\{he, she, they\} said to the } C_{\bar{g}}. 
\end{split}
\end{equation}
where I'm+\{he, she\} combinations produce gender-determined referents, and you're+\{he, she, they\} and I'm+they combinations produce gender-ambiguous referents. There are 22 character pairs, 2 character orders, 2 referent pronoun variants, and 3 speaker pronoun variants, for a total of 264 base templates. Each base template is paired with 4 unique adjectives sampled from the full adjective set, for a total of $N = 1056 = 22\times2\times2\times3\times4$ source sentences (352 determined and 704 ambiguous).

In the two-way conversation template, examples are of the form:
\begin{equation}
\label{eqn:char_both}
\begin{split}
    & \text{The } C_g \text{ smiled. ``I think I'm } A_1 \text{ and you're } A_2\text{,'' } \\
    & \text{they said.} \\
    & \text{The } C_{\bar{g}} \text{ laughed back. ``No, you're } A_3 \text{, but I'm } A_4 \text{,''} \\
    & \text{they replied.} 
\end{split}
\end{equation}
such that the gender of all adjective referents is ambiguous. To observe how the system handles repeated adjectives in the input that refer to different entities, 4 adjective equality variations are used:
\begin{equation}
\label{eqn:a_tuples}
\begin{split}
    & (A_1, A_2, A_3, A_4) \\
    & (A_1, A_2, A_2, A_4) \\
    & (A_1, A_2, A_3, A_1) \\
    & (A_1, A_2, A_2, A_1).
\end{split}    
\end{equation}
There are 22 character pairs, 2 character orders, and 4 adjective equality patterns, for a total of 176 base templates. For each base template, 5 unique adjective tuples $(A_1, A_2, A_3, A_4)$ are sampled from the full adjective set, for a total of $N = 880 = 22 \times 2 \times 4 \times 5$ source sentences. Note that each source sentence provides 4 adjective agreement samples. 

\subsection{Structure Templates}
The structure templates do not include any gender-stereotyped variables, and instead focus on structural variables in dialogue settings between two speakers. There are two template styles: one where all adjectives refer to the same entity, and one where both characters are referenced in equal measure. Both template styles have variations in the provided gender context: two speakers of known gender, such that each adjective's correct gender agreement is always determined, or one known gender and one unknown gender (first-person), such that the adjective's gender is either ambiguous or determined depending on the referent.  

The first template style with complete gender context:
\begin{equation}
    \begin{split}
        & \text{The \{woman, man\} smiled. ``I think \{I'm, you're\} } \\
        & A_1 \text{,'' \{she, he\} said.} \\
        & \text{\{He, She\} laughed back. ``No, [\{you're, I'm\} not } \\
        & A_1 \text{, but] \{you are, I am\} } A_2 \text{,'' \{he, she\} replied.}
    \end{split}
\end{equation}
where the text contained by $[...]$ denotes an optional chaining effect on $A_1$. There are 4 gender combinations for the two characters $\bigl((M, M), (F, F), (F, M), (M, F)\bigr)$, 2 pronoun referent variations (I, you), and 2 chaining variants (present or not), for 16 base templates. For each base template, 60 unique adjective tuples $(A_1, A_2)$ are sampled from the full adjective set, for a total of $N = 960 = 4 \times 2 \times 2 \times 2 \times 60$ source sentences.

The first template style with partial gender context:
\begin{equation}
    \label{eqn:va}
    \begin{split}
        & \text{\{I, The wo/man\} smiled. ``I think \{I'm, you're\} } \\
        & A_1 \text{,'' \{I, s/he\} said.} \\
        & \text{\{S/he, I\} laughed back. ``No, [\{you're, I'm\} not } \\
        & A_1 \text{, but] \{you are, I am\} } A_2 \text{,'' \{s/he, I\} replied.}
    \end{split}
\end{equation}
As above, there are 4 gender combinations $\bigl((M, ?), (F, ?), (?, M), (?, F)\bigr)$, 2 pronoun referent variations, 2 chaining variations, and 60 unique adjective tuples, for a total of $N = 960$ source sentences. The structure variables split this subset in half between ambiguous and determined cases. When the unknown gender (first-person speaker, I) appears first and the first pronoun referent is ``I'', or the known gender speaker appears first and the first person referent is ``you'', all adjectives are gender-ambiguous ($n = 480$ source sentences, $n = 1200$ adjective instances). Otherwise, all adjectives are gender-determined ($n = 480$ source sentences, $n = 1200$ adjective instances). Note that each source sentence contains 2-3 adjective instances, depending on whether the optional chaining effect is included.   

The second template style with complete gender context:
\begin{equation}
    \begin{split}
    & \text{The \{man, woman\} smiled. ``I think I'm } A_1 \text{ and} \\
    & \text{you're } A_2 \text{,'' \{he, she\} said.} \\
    & \text{\{He, She\} laughed back. ``No, you're } A_3 \text{, but I'm } \\
    &  A_4 \text{,'' \{he, she\} replied.} 
    \end{split}
\end{equation}
where there are 4 possible gender combinations, 4 adjective equality patterns as described by equation (\ref{eqn:a_tuples}), and 60 unique adjective tuples $(A_1, A_2, A_3, A_4)$, for a total of $N = 960$ source sentences with 4 determined adjective instances each. 

The second template style with partial gender context:
\begin{equation}
    \begin{split}
    & \text{\{I, The wo/man\} smiled. ``I think I'm } A_1 \text{ and} \\
    & \text{you're } A_2 \text{,'' \{I, s/he\} said.} \\
    & \text{\{S/He, I\} laughed back. ``No, you're } A_3 \text{, but I'm } \\
    &  A_4 \text{,'' \{s/he, I\} replied.} 
    \end{split}
\end{equation}
where again there are 4 possible gender combinations, 4 adjective equality patterns, and 60 unique adjective tuples, for a total of $N = 960$ source sentences. As with template (\ref{eqn:va}), adjectives in this subset are split evenly between ambiguous and determined cases. However, unlike (\ref{eqn:va}), both ambiguous and determined adjectives appear together (equally) in the same source passage. Note that all adjective positions are split evenly between determined and ambiguous cases, as determined by the variable position of the speakers. 

\section{Results for All Systems}
\label{sec:appendix_all_results}

\begin{table*}
\centering
\begin{tabular}{l c c c c c c c}
\hline
\textbf{System} & \textbf{$F$} & \textbf{$M$} & \textbf{$F_{M}$} & \textbf{$M_{M}$} & \textbf{$F_{F}$} & \textbf{$M_{F}$} & \textbf{$\Delta M_{M-F}$} \\
\hline
Aya23	&	0.30308	&	0.69692	&	0.14804	&	0.85196	&	0.53892	&	0.46108	&	0.39088	\\
Claude-3.5	&	0.08439	&	0.91561	&	0.00000	&	1.00000	&	0.14930	&	0.85070	&	0.14930	\\
CommandR-plus	&	0.37172	&	0.62828	&	0.19682	&	0.80318	&	0.51624	&	0.48376	&	0.31942	\\
Dubformer	&	0.08120	&	0.91880	&	0.06135	&	0.93865	&	0.10631	&	0.89369	&	0.04496	\\
GPT-4	&	0.01125	&	0.98875	&	0.00000	&	1.00000	&	0.02308	&	0.97692	&	0.02308	\\
IKUN	&	0.60329	&	0.39671	&	0.49211	&	0.50789	&	0.61719	&	0.38281	&	0.12508	\\
IKUN-C	&	0.49174	&	0.50826	&	0.47581	&	0.52419	&	0.45274	&	0.54726	&	-0.02306	\\
IOL-Research	&	0.08787	&	0.91213	&	0.03196	&	0.96804	&	0.12500	&	0.87500	&	0.09304	\\
Llama3-70B	&	0.03901	&	0.96099	&	0.00000	&	1.00000	&	0.08274	&	0.91726	&	0.08274	\\ 
MSLC	&	0.14011	&	0.85989	&	0.15891	&	0.84109	&	0.19874	&	0.80126	&	0.03983	\\
ONLINE-A	&	0.09344	&	0.90656	&	0.05932	&	0.94068	&	0.11059	&	0.88941	&	0.05126	\\
ONLINE-B	&	0.08717	&	0.91283	&	0.06970	&	0.93030	&	0.09690	&	0.90310	&	0.02721	\\
ONLINE-G	&	0.14204	&	0.85796	&	0.14241	&	0.85759	&	0.16289	&	0.83711	&	0.02048	\\
ONLINE-W	&	0.26113	&	0.73887	&	0.08274	&	0.91726	&	0.49821	&	0.50179	&	0.41548	\\
TranssionMT	&	0.10359	&	0.89641	&	0.10000	&	0.90000	&	0.14516	&	0.85484	&	0.04516	\\
Unbabel-Tower70B	&	0.32142	&	0.67858	&	0.17822	&	0.82178	&	0.42691	&	0.57309	&	0.24869	\\
\hline
\end{tabular}
    \caption{The proportion of adjectives with male ($M$) and female ($F$) agreement on the \textbf{Stereo-Adverb} test suite subset for all systems (English to \textbf{Spanish}). All adjectives self-refer to the speaker of unknown gender. In affected systems, the use of a male-stereotyped adverb to control the manner of speaking increases the use of male adjectives compared to female adjectives (see subscript $M$ denoting the use of male-stereotyped adverbs), and vice versa (see subscript $F$ denoting the use of female-stereotyped adverbs). The baseline, non-subscripted results refer to the proportions of male and female adjective use in the absence of any adverb. The overall strength of the adverb effect can be captured by the difference $\Delta M_{M-F}$.}
    \label{tab:all_adverbs_es}
\end{table*}

\begin{table*}
\centering
\begin{tabular}{l c c c c c c c}
\hline
\textbf{System} & \textbf{$F$} & \textbf{$M$} & \textbf{$F_{M}$} & \textbf{$M_{M}$} & \textbf{$F_{F}$} & \textbf{$M_{F}$} & \textbf{$\Delta M_{M-F}$} \\
\hline
Aya23	&	0.36944	&	0.63056	&	0.18750	&	0.81250	&	0.61244	&	0.38756	&	0.42494	\\
CUNI-DocTransformer	&	0.36044	&	0.63956	&	0.27865	&	0.72135	&	0.52107	&	0.47893	&	0.24241	\\
CUNI-GA	&	0.41955	&	0.58045	&	0.35779	&	0.64221	&	0.47974	&	0.52026	&	0.12195	\\
CUNI-MH	&	0.70343	&	0.29657	&	0.37886	&	0.62114	&	0.95018	&	0.04982	&	0.57132	\\
CUNI-Transformer	&	0.40925	&	0.59075	&	0.37895	&	0.62105	&	0.42209	&	0.57791	&	0.04314	\\
Claude-3.5	&	0.19281	&	0.80719	&	0.00769	&	0.99231	&	0.37334	&	0.62666	&	0.36564	\\
CommandR-plus	&	0.33985	&	0.66015	&	0.11950	&	0.88050	&	0.55371	&	0.44629	&	0.43421	\\
GPT-4	&	0.05730	&	0.94270	&	0.00000	&	1.00000	&	0.11644	&	0.88356	&	0.11644	\\
IKUN	&	0.26889	&	0.73111	&	0.14492	&	0.85508	&	0.32364	&	0.67636	&	0.17872	\\
IKUN-C	&	0.33780	&	0.66220	&	0.26528	&	0.73472	&	0.38904	&	0.61096	&	0.12376	\\
IOL-Research	&	0.04607	&	0.95393	&	0.00000	&	1.00000	&	0.10601	&	0.89399	&	0.10601	\\
Llama3-70B	&	0.02378	&	0.97622	&	0.00000	&	1.00000	&	0.05802	&	0.94198	&	0.05802	\\
NVIDIA-NeMo	&	0.30920	&	0.69080	&	0.31792	&	0.68208	&	0.31086	&	0.68914	&	-0.00706	\\
ONLINE-A	&	0.97638	&	0.02362	&	0.98437	&	0.01563	&	1.00000	&	0.00000	&	0.01563	\\
ONLINE-B	&	0.09241	&	0.90759	&	0.09440	&	0.90560	&	0.11212	&	0.88788	&	0.01771	\\
ONLINE-G	&	0.03956	&	0.96044	&	0.03883	&	0.96117	&	0.03196	&	0.96804	&	-0.00687	\\
ONLINE-W	&	0.59098	&	0.40902	&	0.38679	&	0.61321	&	0.88381	&	0.11619	&	0.49703	\\
SCIR-MT	&	0.21048	&	0.78952	&	0.12845	&	0.87155	&	0.33493	&	0.66507	&	0.20648	\\
TranssionMT	&	0.85793	&	0.14207	&	0.78136	&	0.21864	&	0.91884	&	0.08116	&	0.13747	\\
Unbabel-Tower70B	&	0.38241	&	0.61759	&	0.22141	&	0.77859	&	0.53240	&	0.46760	&	0.31100	\\
\hline
\end{tabular}
    \caption{The proportion of adjectives with male ($M$) and female ($F$) agreement on the \textbf{Stereo-Adverb} test suite subset for all systems (English to \textbf{Czech}).}
    \label{tab:all_adverbs_cz}
\end{table*}

\begin{table*}
\centering
\begin{tabular}{l c c c c c c c}
\hline
\textbf{System} & \textbf{$F$} & \textbf{$M$} & \textbf{$F_{M}$} & \textbf{$M_{M}$} & \textbf{$F_{F}$} & \textbf{$M_{F}$} & \textbf{$\Delta M_{M-F}$} \\
\hline
AMI	&	0.06617	&	0.93383	&	0.04315	&	0.95685	&	0.05593	&	0.94407	&	0.01279	\\
Aya23	&	0.25779	&	0.74221	&	0.13023	&	0.86977	&	0.32917	&	0.67083	&	0.19893	\\
Claude-3.5	&	0.12794	&	0.87206	&	0.01019	&	0.98981	&	0.18941	&	0.81059	&	0.17922	\\
Dubformer	&	0.13975	&	0.86025	&	0.17119	&	0.82881	&	0.13380	&	0.86620	&	-0.03739	\\
GPT-4	&	0.48465	&	0.51535	&	0.36143	&	0.63857	&	0.71391	&	0.28609	&	0.35248	\\
IKUN	&	0.74118	&	0.25882	&	0.61495	&	0.38505	&	0.81479	&	0.18521	&	0.19984	\\
IKUN-C	&	0.42608	&	0.57392	&	0.35849	&	0.64151	&	0.46387	&	0.53613	&	0.10538	\\
IOL-Research	&	0.19206	&	0.80794	&	0.13012	&	0.86988	&	0.25000	&	0.75000	&	0.11988	\\
Llama3-70B	&	0.17153	&	0.82847	&	0.07006	&	0.92994	&	0.24176	&	0.75824	&	0.17170	\\
ONLINE-A	&	0.09271	&	0.90729	&	0.09124	&	0.90876	&	0.09357	&	0.90643	&	0.00234	\\
ONLINE-B	&	0.20944	&	0.79056	&	0.23913	&	0.76087	&	0.18975	&	0.81025	&	-0.04938	\\
ONLINE-G	&	0.14517	&	0.85483	&	0.15385	&	0.84615	&	0.16008	&	0.83992	&	0.00623	\\
TranssionMT	&	0.23640	&	0.76360	&	0.26056	&	0.73944	&	0.25143	&	0.74857	&	-0.00913	\\
Unbabel-Tower70B	&	0.26414	&	0.73586	&	0.11700	&	0.88300	&	0.44016	&	0.55984	&	0.32316	\\
\hline
\end{tabular}
    \caption{The proportion of adjectives with male ($M$) and female ($F$) agreement on the \textbf{Stereo-Adverb} test suite subset for all systems (English to \textbf{Icelandic}).}
    \label{tab:all_adverbs_is}
\end{table*}

\begin{table*}
\centering
\begin{tabular}{l l l l}
\hline
\textbf{Variable} & \textbf{ONLINE-W} & \textbf{Aya23} & \textbf{CommandR-plus} \\
\hline
Intercept	&	$(-1.73,3.7E-07^{***})$	&	$(-1.07,1.1E-04^{***})$	&	$(-0.71,7.6E-03^{**})$	\\
Adj Stereo(M)	&	$(-0.89,1.6E-02^*)$	&	$(-0.15,6.5E-01)$	&	$(-0.82,7.8E-03^{**})$	\\
Adj Stereo(F)	&	$(1.96,4.8E-15^{***})$	&	$(0.48,2.1E-02^*)$	&	$(0.50,1.2E-02^*)$	\\
Adj Sentiment(neg)	&	$(-0.44,3.6E-02^*)$	&	$(0.35,5.6E-02)$	&	$(0.29,9.6E-02)$	\\
Adj Type(appearance)	&	$(0.63,3.1E-03^{**})$	&	$(0.31,9.5E-02)$	&	$(0.52,3.5E-03^{**})$	\\
\textbf{Adv Stereo(M)}	&	$(-0.42,2.0E-01)$	&	$(-0.99,4.4E-04^{***})$	&	$(-0.87,8.5E-04^{***})$	\\
\textbf{Adv Stereo(F)}	&	$(1.61,4.8E-07^{***})$	&	$(0.65,1.0E-02^*)$	&	$(0.54,3.0E-02^*)$	\\
\hline
\end{tabular}
    \caption{Stereotypical manner of speaking (adverb) regression analysis for the most affected systems (\textbf{Spanish}), displayed as (coefficient value, p-value). The variable to predict is the binary adjective declension choice, where feminine adjectives are the positive class, such that negative coefficient values indicate a greater probability of $M$, and positive coefficient values indicate a greater probability of $F$. Strong negative intercepts indicate the default male baseline exhibited by many systems. Internal adjective traits are controlled by stereotype variables (e.g. Stereo(M) is expected to increase the probability of $M$), the sentiment (here negative as opposed to positive), and type (here appearance as opposed to character). For example, if the adjective is the appearance type, the results show that systems ONLINE-W and CommandR-plus are more likely to choose an $F$-adjective, controlling for all other variables. Here we see the adverb variables are strong in their expected directions, and significant.}
    \label{tab:adverb_reg_es}
\end{table*}

\begin{table*}
\centering
\begin{tabular}{l l l l}
\hline
\textbf{Variable} & \textbf{CUNI-MH} & \textbf{ONLINE-W} & \textbf{CommandR-plus} \\
\hline
Intercept	&	$(2.45,1.2E-07^{***})$	&	$(-0.66,1.1E-02^*)$	&	$(-0.84,1.3E-03^{**})$	\\
Adj Stereo(M)	&	$(-0.96,8.1E-03^{**})$	&	$(-0.76,1.5E-02^*)$	&	$(-0.60,6.3E-02)$	\\
Adj Stereo(F)	&	$(1.04,1.6E-04^{***})$	&	$(0.95,2.7E-05^{***})$	&	$(1.30,5.1E-09^{***})$	\\
Adj Sentiment(neg)	&	$(0.13,5.7E-01)$	&	$(-0.47,1.2E-02^*)$	&	$(-0.19,3.2E-01)$	\\
Adj Type(appearance)	&	$(0.57,1.4E-02^*)$	&	$(1.04,8.9E-08^{***})$	&	$(0.94,1.1E-06^{***})$	\\
\textbf{Adv Stereo(M)}	&	$(-3.26,7.0E-13^{***})$	&	$(0.01,9.6E-01)$	&	$(-1.59,1.2E-08^{***})$	\\
\textbf{Adv Stereo(F)}	&	$(0.05,9.3E-01)$	&	$(2.26,5.3E-16^{***})$	&	$(0.68,6.3E-03^{**})$	\\
\hline
\end{tabular}
    \caption{Stereotypical manner of speaking (adverb) regression analysis for the most affected systems (\textbf{Czech}).}
    \label{tab:adverb_reg_cs}
\end{table*}

\begin{table*}
\centering
\begin{tabular}{l l l l}
\hline
\textbf{Variable} & \textbf{GPT-4} & \textbf{Unbabel-Tower70B} & \textbf{IKUN} \\
\hline
Intercept	&	$(0.52,4.7E-02^*)$	&	$(-0.91,1.4E-03^{**})$	&	$(1.31,1.6E-05^{***})$	\\
Adj Stereo(M)	&	$(-1.51,4.7E-07^{***})$	&	$(-0.72,2.9E-02^*)$	&	$(-0.78,3.6E-03^{**})$	\\
Adj Stereo(F)	&	$(0.75,2.6E-04^{***})$	&	$(0.89,1.4E-05^{***})$	&	$(0.62,1.1E-02^*)$	\\
Adj Sentiment(neg)	&	$(-0.22,2.1E-01)$	&	$(-0.49,9.7E-03^{**})$	&	$(-0.01,9.6E-01)$	\\
Adj Type(appearance)	&	$(0.28,1.1E-01)$	&	$(0.03,8.7E-01)$	&	$(-0.08,6.5E-01)$	\\
\textbf{Adv Stereo(M)}	&	$(-0.98,1.5E-04^{***})$	&	$(-0.54,5.7E-02)$	&	$(-0.64,2.6E-02^*)$	\\
\textbf{Adv Stereo(F)}	&	$(0.31,2.3E-01)$	&	$(0.70,9.0E-03^{**})$	&	$(0.07,8.2E-01)$	\\
\hline
\end{tabular}
    \caption{Stereotypical manner of speaking (adverb) regression analysis for the most affected systems (\textbf{Icelandic}).}
    \label{tab:adverb_reg_is}
\end{table*}

\begin{table*}
\centering
\begin{tabular}{l c c c c c c c}
\hline
\textbf{System} & \textbf{$F_{M}$} & \textbf{$M_{M}$} & \textbf{$F_{F}$} & \textbf{$M_{F}$} & \textbf{$\Delta M_{M-F}$} \\
\hline
Aya23	&	0.122065	&	0.877935	&	0.429200	&	0.570800	&	0.307135	\\
Claude-3.5	&	0.000000	&	1.000000	&	0.390775	&	0.609225	&	0.390775	\\
CommandR-plus	&	0.011598	&	0.988402	&	0.401382	&	0.598618	&	0.389784	\\
Dubformer	&	0.002979	&	0.997021	&	0.128387	&	0.871613	&	0.125408	\\
GPT-4	&	0.000000	&	1.000000	&	0.273960	&	0.726040	&	0.273960	\\
IKUN	&	0.165404	&	0.834596	&	0.363451	&	0.636549	&	0.198047	\\
IKUN-C	&	0.287342	&	0.712658	&	0.418834	&	0.581166	&	0.131493	\\
IOL-Research	&	0.003182	&	0.996818	&	0.245182	&	0.754818	&	0.242001	\\
Llama3-70B	&	0.000000	&	1.000000	&	0.261807	&	0.738193	&	0.261807	\\
MSLC	&	0.263380	&	0.736620	&	0.217680	&	0.782320	&	-0.045700	\\
ONLINE-A	&	0.023395	&	0.976605	&	0.109371	&	0.890629	&	0.085976	\\
ONLINE-B	&	0.038452	&	0.961548	&	0.084521	&	0.915479	&	0.046069	\\
ONLINE-G	&	0.054620	&	0.945380	&	0.042830	&	0.957170	&	-0.011790	\\
ONLINE-W	&	0.055130	&	0.944870	&	0.175764	&	0.824236	&	0.120634	\\
TranssionMT	&	0.020904	&	0.979096	&	0.100892	&	0.899108	&	0.079988	\\
Unbabel-Tower70B	&	0.057582	&	0.942418	&	0.359154	&	0.640846	&	0.301572	\\
\hline
\end{tabular}
    \caption{The proportion of adjectives with male ($M$) and female ($F$) agreement on the \textbf{Stereo-Character-Amb} test suite subset for unknown gender cases for all systems (English to \textbf{Spanish}). All adjectives refer to either the speaker or the listener which have been introduced as gender-stereotyped characters. The subscripts denote the gender stereotype label. In affected systems, adjectives that refer to a $M$-stereotyped character (subscript $M$) are more likely to be translated with a male declension, and vice versa for $F$-stereotyped characters. The overall strength of the character description effect can be captured by the difference $\Delta M_{M-F}$.}
    \label{tab:all_char_amb_es}
\end{table*}

\begin{table*}
\centering
\begin{tabular}{l c c c c c c c}
\hline
\textbf{System} & \textbf{$F_{M}$} & \textbf{$M_{M}$} & \textbf{$F_{F}$} & \textbf{$M_{F}$} & \textbf{$\Delta M_{M-F}$} \\
\hline
Aya23	&	0.157011	&	0.842989	&	0.490925	&	0.509075	&	0.333913	\\
CUNI-DocTransformer	&	0.194503	&	0.805497	&	0.304000	&	0.696000	&	0.109497	\\
CUNI-GA	&	0.237413	&	0.762587	&	0.328323	&	0.671677	&	0.090910	\\
CUNI-MH	&	0.184362	&	0.815638	&	0.649108	&	0.350892	&	0.464746	\\
CUNI-Transformer	&	0.279416	&	0.720584	&	0.329398	&	0.670602	&	0.049982	\\
Claude-3.5	&	0.020730	&	0.979270	&	0.424670	&	0.575330	&	0.403940	\\
CommandR-plus	&	0.020468	&	0.979532	&	0.354000	&	0.646000	&	0.333532	\\
GPT-4	&	0.005208	&	0.994792	&	0.348186	&	0.651814	&	0.342978	\\
IKUN	&	0.125093	&	0.874907	&	0.402927	&	0.597073	&	0.277834	\\
IKUN-C	&	0.273129	&	0.726871	&	0.528897	&	0.471103	&	0.255767	\\
IOL-Research	&	0.037213	&	0.962787	&	0.350441	&	0.649559	&	0.313228	\\
Llama3-70B	&	0.002959	&	0.997041	&	0.193021	&	0.806979	&	0.190062	\\
NVIDIA-NeMo	&	0.127240	&	0.872760	&	0.261404	&	0.738596	&	0.134165	\\
ONLINE-A	&	0.140739	&	0.859261	&	0.223089	&	0.776911	&	0.082350	\\
ONLINE-B	&	0.050206	&	0.949794	&	0.097500	&	0.902500	&	0.047294	\\
ONLINE-G	&	0.063216	&	0.936784	&	0.070601	&	0.929399	&	0.007385	\\
ONLINE-W	&	0.085828	&	0.914172	&	0.458372	&	0.541628	&	0.372544	\\
SCIR-MT	&	0.082652	&	0.917348	&	0.277897	&	0.722103	&	0.195246	\\
TranssionMT	&	0.108245	&	0.891755	&	0.178112	&	0.821888	&	0.069868	\\
Unbabel-Tower70B	&	0.048333	&	0.951667	&	0.381000	&	0.619000	&	0.332667	\\
\hline
\end{tabular}
    \caption{The proportion of adjectives with male ($M$) and female ($F$) agreement on the \textbf{Stereo-Character-Amb} test suite subset for unknown gender cases for all systems (English to \textbf{Czech}).}
    \label{tab:all_char_amb_cz}
\end{table*}

\begin{table*}
\centering
\begin{tabular}{l c c c c c c c}
\hline
\textbf{System} & \textbf{$F_{M}$} & \textbf{$M_{M}$} & \textbf{$F_{F}$} & \textbf{$M_{F}$} & \textbf{$\Delta M_{M-F}$} \\
\hline
AMI	&	0.106313	&	0.893687	&	0.077960	&	0.922040	&	-0.028353	\\
Aya23	&	0.234091	&	0.765909	&	0.450333	&	0.549667	&	0.216242	\\
Claude-3.5	&	0.005272	&	0.994728	&	0.434516	&	0.565484	&	0.429244	\\
Dubformer	&	0.092273	&	0.907727	&	0.139669	&	0.860331	&	0.047396	\\
GPT-4	&	0.159235	&	0.840765	&	0.477160	&	0.522840	&	0.317924	\\
IKUN	&	0.282857	&	0.717143	&	0.555590	&	0.444410	&	0.272733	\\
IKUN-C	&	0.273029	&	0.726971	&	0.378756	&	0.621244	&	0.105728	\\
IOL-Research	&	0.002394	&	0.997606	&	0.126020	&	0.873980	&	0.123627	\\
Llama3-70B	&	0.065437	&	0.934563	&	0.235696	&	0.764304	&	0.170259	\\
ONLINE-A	&	0.040660	&	0.959340	&	0.031744	&	0.968256	&	-0.008916	\\
ONLINE-B	&	0.122783	&	0.877217	&	0.087892	&	0.912108	&	-0.034891	\\
ONLINE-G	&	0.089511	&	0.910489	&	0.048492	&	0.951508	&	-0.041019	\\
TranssionMT	&	0.103972	&	0.896028	&	0.100703	&	0.899297	&	-0.003269	\\
Unbabel-Tower70B	&	0.063657	&	0.936343	&	0.293988	&	0.706012	&	0.230331	\\
\hline
\end{tabular}
    \caption{The proportion of adjectives with male ($M$) and female ($F$) agreement on the \textbf{Stereo-Character-Amb} test suite subset for unknown gender cases for all systems (English to \textbf{Icelandic}).}
    \label{tab:all_char_amb_is}
\end{table*}

\begin{table*}
\centering
\begin{tabular}{l l l l}
\hline
\textbf{Variable} & \textbf{Claude-3.5} & \textbf{CommandR-plus} & \textbf{Aya23} \\
\hline
Intercept	&	$(-6.83,2.7E-11^{***})$	&	$(-3.89,5.5E-34^{***})$	&	$(-0.99,3.1E-07^{***})$	\\
Adj Stereo(M)	&	$(0.87,8.2E-02)$	&	$(0.29,5.4E-01)$	&	$(-0.13,7.9E-01)$	\\
Adj Stereo(F)	&	$(0.85,2.7E-06^{***})$	&	$(-0.29,1.3E-01)$	&	$(0.15,2.5E-01)$	\\
Adj Sentiment(neg)	&	$(-0.56,1.6E-04^{***})$	&	$(0.04,7.4E-01)$	&	$(-0.86,4.9E-16^{***})$	\\
Adj Type(appearance)	&	$(0.01,9.7E-01)$	&	$(0.67,3.6E-04^{***})$	&	$(0.06,7.0E-01)$	\\
\textbf{Character Stereo(F)}	&	$(6.59,5.0E-11^{***})$	&	$(3.56,2.4E-43^{***})$	&	$(1.41,9.5E-43^{***})$	\\
\hline
\end{tabular}
    \caption{Stereotypical character description regression analysis for the most affected systems (\textbf{Spanish}). Internal adjective variables are defined as above (see Table \ref{tab:adverb_reg_es}). Here \textbf{Character Stereo(F)} denotes a binary variable equal to 1 when the character description is stereotypically female, and equal to 0 when the character description is stereotypically male. As shown, coefficient values for \textbf{Character Stereo(F)} are significant and in the expected direction (positive, indicating an increased likelihood of a $F$-adjective), and much stronger than the internal variables.}
    \label{tab:char_reg_es}
\end{table*}

\begin{table*}
\centering
\begin{tabular}{l l l l}
\hline
\textbf{Variable} & \textbf{CUNI-MH} & \textbf{Claude-3.5} & \textbf{ONLINE-W} \\
\hline
Intercept	&	$(-1.38,1.5E-16^{***})$	&	$(-4.93,1.2E-50^{***})$	&	$(-1.36,9.4E-13^{***})$	\\
Adj Stereo(M)	&	$(-0.53,3.6E-03^{**})$	&	$(-0.75,8.6E-03^{**})$	&	$(-1.01,6.7E-02)$	\\
Adj Stereo(F)	&	$(1.58,1.7E-37^{***})$	&	$(0.17,3.0E-01)$	&	$(0.01,9.5E-01)$	\\
Adj Sentiment(neg)	&	$(-0.43,9.8E-08^{***})$	&	$(-0.31,6.9E-03^{**})$	&	$(-0.64,2.1E-11^{***})$	\\
Adj Type(appearance)	&	$(-0.78,5.6E-09^{***})$	&	$(0.20,3.1E-01)$	&	$(1.34,2.5E-26^{***})$	\\
\textbf{Character Stereo(F)}	&	$(1.66,1.1E-93^{***})$	&	$(4.25,4.4E-52^{***})$	&	$(2.20,1.1E-76^{***})$	\\
\hline
\end{tabular}
    \caption{Stereotypical character description regression analysis for the most affected systems (\textbf{Czech}).}
    \label{tab:char_reg_cs}
\end{table*}

\begin{table*}
\centering
\begin{tabular}{l l l l}
\hline
\textbf{Variable} & \textbf{Claude-3.5} & \textbf{GPT-4} & \textbf{IKUN} \\
\hline
Intercept	&	$(-4.09,9.7E-45^{***})$	&	$(-1.93,3.0E-31^{***})$	&	$(-1.13,5.6E-14^{***})$	\\
Adj Stereo(M)	&	$(-1.55,1.3E-06^{***})$	&	$(-0.18,3.3E-01)$	&	$(0.12,6.7E-01)$	\\
Adj Stereo(F)	&	$(-0.45,1.5E-02^*)$	&	$(0.04,7.9E-01)$	&	$(0.48,2.1E-04^{***})$	\\
Adj Sentiment(neg)	&	$(0.91,2.1E-15^{***})$	&	$(0.34,6.9E-05^{***})$	&	$(0.80,2.3E-24^{***})$	\\
Adj Type(appearance)	&	$(0.03,8.8E-01)$	&	$(-0.03,8.3E-01)$	&	$(0.69,1.8E-09^{***})$	\\
Character Stereo(F)	&	$(3.62,1.3E-64^{***})$	&	$(1.16,1.5E-44^{***})$	&	$(0.60,4.7E-15^{***})$	\\
\hline
\end{tabular}
    \caption{Stereotypical character description regression analysis for the most affected systems (\textbf{Icelandic}).}
    \label{tab:char_reg_is}
\end{table*}

\begin{table*}
\centering
\begin{tabular}{l c c c}
\hline
\textbf{System} & \textbf{Accuracy (PRO)} & \textbf{Accuracy (ANTI)} & \textbf{$\Delta($PRO, ANTI$)$} \\
\hline
Aya23	&	1.000	&	0.655000	&	0.345000 \\
Claude-3.5	&	1.000	&	0.742500	&	0.257500 \\
CommandR-plus	&	1.000	&	0.950167	&	0.049833 \\
Dubformer	&	0.965	&	0.776000	&	0.189000 \\
GPT-4	&	0.990	&	0.527500	&	0.462500 \\
IKUN	&	0.975	&	0.701667	&	0.273333 \\
IKUN-C	&	0.910	&	0.887167	&	0.022833 \\
IOL-Research	&	0.990	&	0.963500	&	0.026500 \\
Llama3-70B	&	1.000	&	0.862500	&	0.137500 \\
MSLC	&	0.935	&	0.870500	&	0.064500 \\
ONLINE-A	&	0.970	&	0.990667	&	-0.020667 \\
ONLINE-B	&	0.980	&	0.996000	&	-0.016000 \\
ONLINE-G	&	1.000	&	0.996333	&	0.003667 \\
ONLINE-W	&	0.985	&	0.414333	&	0.570667 \\
TranssionMT	&	0.985	&	0.990167	&	-0.005167 \\
Unbabel-Tower70B	&	0.995	&	0.970833	&	0.024167 \\
\hline
\end{tabular}
    \caption{The accuracy in gender-adjective agreement on the \textbf{Stereo-Character-Det} test suite subset for known gender cases for all systems (English to \textbf{Spanish}). The test subset is further partitioned into cases that align with a stereotype (PRO) and cases that oppose a stereotype (ANTI). Accuracy is consistently high on the PRO subset, and drops significantly in the challenge setting for some translation systems, indicating that stereotype effects persist in the presence of correct and unambiguous gender context.}
    \label{tab:all_char_det_es}
\end{table*}

\begin{table*}
\centering
\begin{tabular}{l c c c}
\hline
\textbf{System} & \textbf{Accuracy (PRO)} & \textbf{Accuracy (ANTI)} & \textbf{$\Delta($PRO, ANTI$)$} \\
\hline
Aya23	&	0.985	&	0.539833	&	0.445167	\\
CUNI-DocTransformer	&	1.000	&	0.993500	&	0.006500	\\
CUNI-GA	&	0.995	&	0.991667	&	0.003333	\\
CUNI-MH	&	1.000	&	0.995500	&	0.004500	\\
CUNI-Transformer	&	1.000	&	0.994167	&	0.005833	\\
Claude-3.5	&	1.000	&	0.878167	&	0.121833	\\
CommandR-plus	&	0.985	&	0.912333	&	0.072667	\\
GPT-4	&	1.000	&	0.807833	&	0.192167	\\
IKUN	&	0.935	&	0.814000	&	0.121000	\\
IKUN-C	&	0.995	&	0.930000	&	0.065000	\\
IOL-Research	&	1.000	&	0.878000	&	0.122000	\\
Llama3-70B	&	1.000	&	0.640000	&	0.360000	\\
NVIDIA-NeMo	&	1.000	&	0.993333	&	0.006667	\\
ONLINE-A	&	1.000	&	0.996333	&	0.003667	\\
ONLINE-B	&	1.000	&	0.996667	&	0.003333	\\
ONLINE-G	&	1.000	&	1.000000	&	0.000000	\\
ONLINE-W	&	1.000	&	0.988667	&	0.011333	\\
SCIR-MT	&	1.000	&	0.871667	&	0.128333	\\
TranssionMT	&	1.000	&	0.994667	&	0.005333	\\
Unbabel-Tower70B	&	1.000	&	0.867000	&	0.133000	\\
\hline
\end{tabular}
    \caption{The accuracy in gender-adjective agreement on the \textbf{Stereo-Character-Det} test suite subset for known gender cases for all systems (English to \textbf{Czech}).}
    \label{tab:all_char_det_cz}
\end{table*}

\begin{table*}
\centering
\begin{tabular}{l c c c}
\hline
\textbf{System} & \textbf{Accuracy (PRO)} & \textbf{Accuracy (ANTI)} & \textbf{$\Delta($PRO, ANTI$)$} \\
\hline
AMI	&	0.990000	&	0.977667	&	0.012333 \\
Aya23	&	0.632833	&	0.765833	&	-0.133000 \\
Claude-3.5	&	0.990000	&	0.900833	&	0.089167 \\
Dubformer	&	0.657560	&	0.564000	&	0.093560 \\
GPT-4	&	0.925000	&	0.832500	&	0.092500 \\
IKUN	&	0.890000	&	0.942167	&	-0.052167 \\
IKUN-C	&	0.975000	&	0.950333	&	0.024667 \\
IOL-Research	&	0.955000	&	0.963167	&	-0.008167 \\
Llama3-70B	&	0.982416	&	0.850000	&	0.132416 \\
ONLINE-A	&	0.885000	&	0.973833	&	-0.088833 \\
ONLINE-B	&	1.000000	&	0.988500	&	0.011500 \\
ONLINE-G	&	0.940000	&	0.830000	&	0.110000 \\
TranssionMT	&	1.000000	&	0.982833	&	0.017167 \\
Unbabel-Tower70B	&	1.000000	&	0.969167	&	0.030833 \\
\hline
\end{tabular}
    \caption{The accuracy in gender-adjective agreement on the \textbf{Stereo-Character-Det} test suite subset for known gender cases for all systems (English to \textbf{Icelandic}).}
    \label{tab:all_char_det_is}
\end{table*}

\begin{table*}
\centering
\begin{tabular}{l c c c c c c c}
\hline
\textbf{System} & \textbf{$F_{M}$} & \textbf{$M_{M}$} & \textbf{$F_{F}$} & \textbf{$M_{F}$} & \textbf{$\Delta M_{M-F}$} \\
\hline
Aya23	&	0.256023	&	0.743977	&	0.345843	&	0.654157	&	0.089820	\\
Claude-3.5	&	0.419257	&	0.580743	&	0.073611	&	0.926389	&	-0.345646	\\
CommandR-plus	&	0.763592	&	0.236408	&	0.425954	&	0.574046	&	-0.337638	\\
Dubformer	&	0.089736	&	0.910264	&	0.110243	&	0.889757	&	0.020507	\\
GPT-4	&	0.090745	&	0.909255	&	0.019064	&	0.980936	&	-0.071682	\\
IKUN	&	0.255602	&	0.744398	&	0.725592	&	0.274408	&	0.469990	\\
IKUN-C	&	0.292295	&	0.707705	&	0.702554	&	0.297446	&	0.410258	\\
IOL-Research	&	0.070826	&	0.929174	&	0.342207	&	0.657793	&	0.271382	\\
Llama3-70B	&	0.105211	&	0.894789	&	0.064873	&	0.935127	&	-0.040338	\\
MSLC	&	0.177329	&	0.822671	&	0.247036	&	0.752964	&	0.069707	\\
ONLINE-A	&	0.049758	&	0.950242	&	0.208951	&	0.791049	&	0.159194	\\
ONLINE-B	&	0.105807	&	0.894193	&	0.141280	&	0.858720	&	0.035473	\\
ONLINE-G	&	0.070760	&	0.929240	&	0.300656	&	0.699344	&	0.229895	\\
ONLINE-W	&	0.261696	&	0.738304	&	0.372914	&	0.627086	&	0.111218	\\
TranssionMT	&	0.091749	&	0.908251	&	0.151915	&	0.848085	&	0.060166	\\
Unbabel-Tower70B	&	0.273571	&	0.726429	&	0.415857	&	0.584143	&	0.142286	\\
\hline
\end{tabular}
    \caption{The proportion of adjectives with male ($M$) and female ($F$) agreement on the \textbf{Structure-Amb} test suite subset for all systems (English to \textbf{Spanish}). All adjectives refer to someone of an unknown gender in conversation with someone of a known gender (where that known gender is denoted by the subscripts). 
    Systems that have an ``opposite'' binary gender bias effect resolve the ambiguous-gender speaker to be opposite to the known speaker (i.e., masculine adjectives increase when the other speaker is female, and vice versa, and the difference $\Delta M_{M-F}$ is strongly positive). Systems with a same-binary gender effect consistently choose adjective forms matching the gender of the other speaker (i.e., $\Delta M_{M-F}$ is strongly negative).}
    \label{tab:all_char_amb}
\end{table*}

\begin{table*}
\centering
\begin{tabular}{l c c c c c c c}
\hline
\textbf{System} & \textbf{$F_{M}$} & \textbf{$M_{M}$} & \textbf{$F_{F}$} & \textbf{$M_{F}$} & \textbf{$\Delta M_{M-F}$} \\
\hline
Aya23	&	0.671295	&	0.328705	&	0.583859	&	0.416141	&	-0.087436	\\
CUNI-DocTransformer	&	0.092894	&	0.907106	&	0.270689	&	0.729311	&	0.177795	\\
CUNI-GA	&	0.455846	&	0.544154	&	0.175642	&	0.824358	&	-0.280204	\\
CUNI-MH	&	0.714519	&	0.285481	&	0.699626	&	0.300374	&	-0.014893	\\
CUNI-Transformer	&	0.454953	&	0.545047	&	0.173387	&	0.826613	&	-0.281566	\\
Claude-3.5	&	0.380231	&	0.619769	&	0.039362	&	0.960638	&	-0.340869	\\
CommandR-plus	&	0.661912	&	0.338088	&	0.139604	&	0.860396	&	-0.522308	\\
GPT-4	&	0.496953	&	0.503047	&	0.033441	&	0.966559	&	-0.463512	\\
IKUN	&	0.118729	&	0.881271	&	0.322997	&	0.677003	&	0.204268	\\
IKUN-C	&	0.475241	&	0.524759	&	0.808473	&	0.191527	&	0.333232	\\
IOL-Research	&	0.131136	&	0.868864	&	0.053251	&	0.946749	&	-0.077885	\\
Llama3-70B	&	0.041654	&	0.958346	&	0.014288	&	0.985712	&	-0.027366	\\
NVIDIA-NeMo	&	0.027024	&	0.972976	&	0.635958	&	0.364042	&	0.608934	\\
ONLINE-A	&	0.733600	&	0.266400	&	0.381812	&	0.618188	&	-0.351788	\\
ONLINE-B	&	0.058142	&	0.941858	&	0.082191	&	0.917809	&	0.024049	\\
ONLINE-G	&	0.025188	&	0.974812	&	0.192564	&	0.807436	&	0.167376	\\
ONLINE-W	&	0.645775	&	0.354225	&	0.390405	&	0.609595	&	-0.255370	\\
SCIR-MT	&	0.361420	&	0.638580	&	0.440349	&	0.559651	&	0.078929	\\
TranssionMT	&	0.586897	&	0.413103	&	0.326415	&	0.673585	&	-0.260481	\\
Unbabel-Tower70B	&	0.443972	&	0.556028	&	0.323738	&	0.676262	&	-0.120234	\\
\hline
\end{tabular}
    \caption{The proportion of adjectives with male ($M$) and female ($F$) agreement on the \textbf{Structure-Amb} test suite subset for all systems (English to \textbf{Czech}).}
    \label{tab:all_char_amb}
\end{table*}

\begin{table*}
\centering
\begin{tabular}{l c c c c c c c}
\hline
\textbf{System} & \textbf{$F_{M}$} & \textbf{$M_{M}$} & \textbf{$F_{F}$} & \textbf{$M_{F}$} & \textbf{$\Delta M_{M-F}$} \\
\hline
AMI	&	0.078522	&	0.921478	&	0.345131	&	0.654869	&	0.266609	\\
Aya23	&	0.298890	&	0.701110	&	0.269610	&	0.730390	&	-0.029280	\\
Claude-3.5	&	0.561121	&	0.438879	&	0.151645	&	0.848355	&	-0.409476	\\
Dubformer	&	0.087548	&	0.912452	&	0.158890	&	0.841110	&	0.071343	\\
GPT-4	&	0.683264	&	0.316736	&	0.517260	&	0.482740	&	-0.166004	\\
IKUN	&	0.497869	&	0.502131	&	0.862015	&	0.137985	&	0.364145	\\
IKUN-C	&	0.302331	&	0.697669	&	0.695231	&	0.304769	&	0.392900	\\
IOL-Research	&	0.085036	&	0.914964	&	0.236664	&	0.763336	&	0.151628	\\
Llama3-70B	&	0.248973	&	0.751027	&	0.333898	&	0.666102	&	0.084925	\\
ONLINE-A	&	0.035666	&	0.964334	&	0.139427	&	0.860573	&	0.103761	\\
ONLINE-B	&	0.140336	&	0.859664	&	0.256017	&	0.743983	&	0.115682	\\
ONLINE-G	&	0.069463	&	0.930537	&	0.136131	&	0.863869	&	0.066667	\\
TranssionMT	&	0.142135	&	0.857865	&	0.253417	&	0.746583	&	0.111282	\\
Unbabel-Tower70B	&	0.207583	&	0.792417	&	0.407592	&	0.592408	&	0.200009	\\
\hline
\end{tabular}
    \caption{The proportion of adjectives with male ($M$) and female ($F$) agreement on the \textbf{Structure-Amb} test suite subset for all systems (English to \textbf{Icelandic}).}
    \label{tab:all_char_amb}
\end{table*}

\begin{table*}
\centering
\begin{tabular}{l c c c c}
\hline
\textbf{System} & \textbf{Acc (one gender)} & \textbf{Acc (same genders)} & \textbf{Acc (opp genders)} & \textbf{$\Delta($same, opp$)$} \\
\hline
Aya23	&	0.937340	&	0.812066	&	0.930527	&	-0.118461	\\
Claude-3.5	&	0.997078	&	0.923965	&	0.997372	&	-0.073407	\\
CommandR-plus	&	0.987414	&	0.796548	&	0.990717	&	-0.194170	\\
Dubformer	&	0.844990	&	0.790325	&	0.850310	&	-0.059985	\\
GPT-4	&	0.991524	&	0.855742	&	0.992963	&	-0.137221	\\
IKUN	&	0.876698	&	0.835986	&	0.837003	&	-0.001018	\\
IKUN-C	&	0.863909	&	0.838490	&	0.798583	&	0.039907	\\
IOL-Research	&	0.947063	&	0.873722	&	0.906976	&	-0.033254	\\
Llama3-70B	&	0.956589	&	0.805900	&	0.977354	&	-0.171454	\\
MSLC	&	0.611581	&	0.692553	&	0.598783	&	0.093771	\\
ONLINE-A	&	0.734181	&	0.828018	&	0.667730	&	0.160288	\\
ONLINE-B	&	0.727604	&	0.740764	&	0.746103	&	-0.005339	\\
ONLINE-G	&	0.725552	&	0.826803	&	0.624745	&	0.202058	\\
ONLINE-W	&	0.914281	&	0.887881	&	0.919022	&	-0.031141	\\
TranssionMT	&	0.728791	&	0.739865	&	0.748009	&	-0.008144	\\
Unbabel-Tower70B	&	0.924064	&	0.817639	&	0.909270	&	-0.091631	\\
\hline
\end{tabular}
    \caption{The accuracy in gender-adjective agreement on the \textbf{Structure-Det} test suite subset for known gender cases for all systems (English to \textbf{Spanish}). The second speaker in the conversation is either unknown (one gender subset), the same, or opposite to the adjective referent. Systems with an opposite binary gender effect suffer on the same-gender subset such that the difference in accuracy $\Delta($same, opp$) \ll 0$, and systems with a same-gender preference suffer on the opposite-gender subset such that the difference in accuracy $\Delta($same, opp$) \gg 0$.}
    \label{tab:all_char_det_es}
\end{table*}

\begin{table*}
\centering
\begin{tabular}{l c c c c}
\hline
\textbf{System} & \textbf{Acc (one gender)} & \textbf{Acc (same genders)} & \textbf{Acc (opp genders)} & \textbf{$\Delta($same, opp$)$} \\
\hline
Aya23	&	0.965847	&	0.808469	&	0.951880	&	-0.143411	\\
CUNI-DocTransformer	&	0.892850	&	0.896380	&	0.855471	&	0.040909	\\
CUNI-GA	&	0.768732	&	0.601509	&	0.911337	&	-0.309828	\\
CUNI-MH	&	0.928232	&	0.814084	&	0.898555	&	-0.084471	\\
CUNI-Transformer	&	0.769805	&	0.602070	&	0.911349	&	-0.309278	\\
Claude-3.5	&	0.995241	&	0.911843	&	0.999082	&	-0.087239	\\
CommandR-plus	&	0.996232	&	0.739137	&	0.990259	&	-0.251121	\\
GPT-4	&	0.997750	&	0.823093	&	0.989451	&	-0.166358	\\
IKUN	&	0.856785	&	0.812804	&	0.878981	&	-0.066177	\\
IKUN-C	&	0.883146	&	0.867369	&	0.831151	&	0.036219	\\
IOL-Research	&	0.975905	&	0.916801	&	0.957697	&	-0.040897	\\
Llama3-70B	&	0.953773	&	0.827831	&	0.931734	&	-0.103903	\\
NVIDIA-NeMo	&	0.824710	&	0.797210	&	0.704606	&	0.092604	\\
ONLINE-A	&	0.736260	&	0.561771	&	0.926380	&	-0.364608	\\
ONLINE-B	&	0.760273	&	0.752525	&	0.750789	&	0.001737	\\
ONLINE-G	&	0.739938	&	0.812760	&	0.667776	&	0.144984	\\
ONLINE-W	&	0.892118	&	0.828515	&	0.927573	&	-0.099059	\\
SCIR-MT	&	0.916721	&	0.835374	&	0.869839	&	-0.034465	\\
TranssionMT	&	0.742036	&	0.597267	&	0.894969	&	-0.297702	\\
Unbabel-Tower70B	&	0.935530	&	0.863328	&	0.926457	&	-0.063129	\\
\hline
\end{tabular}
    \caption{The accuracy in gender-adjective agreement on the \textbf{Structure-Det} test suite subset for known gender cases for all systems (English to \textbf{Czech}).}
    \label{tab:all_char_det_cs}
\end{table*}

\begin{table*}
\centering
\begin{tabular}{l c c c c}
\hline
\textbf{System} & \textbf{Acc (one gender)} & \textbf{Acc (same genders)} & \textbf{Acc (opp genders)} & \textbf{$\Delta($same, opp$)$} \\
\hline
AMI	&	0.741426	&	0.890035	&	0.606934	&	0.283102	\\
Aya23	&	0.650749	&	0.665003	&	0.681895	&	-0.016892	\\
Claude-3.5	&	0.990550	&	0.948800	&	0.983105	&	-0.034305	\\
Dubformer	&	0.685313	&	0.663091	&	0.701806	&	-0.038716	\\
GPT-4	&	0.923000	&	0.862593	&	0.906645	&	-0.044052	\\
IKUN	&	0.859620	&	0.793285	&	0.788005	&	0.005280	\\
IKUN-C	&	0.860037	&	0.826175	&	0.774361	&	0.051815	\\
IOL-Research	&	0.927880	&	0.879636	&	0.890744	&	-0.011107	\\
Llama3-70B	&	0.863632	&	0.784824	&	0.830711	&	-0.045887	\\
ONLINE-A	&	0.681548	&	0.743801	&	0.602983	&	0.140818	\\
ONLINE-B	&	0.745792	&	0.794195	&	0.697057	&	0.097137	\\
ONLINE-G	&	0.579342	&	0.617988	&	0.546436	&	0.071552	\\
TranssionMT	&	0.747217	&	0.795348	&	0.691245	&	0.104103	\\
Unbabel-Tower70B	&	0.933936	&	0.892148	&	0.916886	&	-0.024738	\\
\hline
\end{tabular}
    \caption{The accuracy in gender-adjective agreement on the \textbf{Structure-Det} test suite subset for known gender cases for all systems (English to \textbf{Icelandic}).}
    \label{tab:all_char_det_is}
\end{table*}

\begin{table*}
\centering
\begin{tabular}{l l l l}
\hline
\textbf{Variable} & \textbf{CommandR-plus} & \textbf{Llama3-70B} & \textbf{GPT-4} \\
\hline
Intercept	&	$(5.31,1.3E-24^{***})$	&	$(3.69,8.8E-41^{***})$	&	$(6.88,1.1E-11^{***})$	\\
True(M)	&	$(-12.55,9.4E-96^{***})$	&	$(-11.53,8.9E-129^{***})$	&	$(-14.32,2.9E-40^{***})$	\\
Adj Stereo(M)	&	$(-0.03,9.0E-01)$	&	$(-0.58,5.8E-03^{**})$	&	$(-0.95,6.5E-05^{***})$	\\
Adj Stereo(F)	&	$(1.26,2.7E-17^{***})$	&	$(1.77,5.6E-24^{***})$	&	$(1.89,7.5E-24^{***})$	\\
Adj Sentiment(neg)	&	$(-0.66,4.6E-07^{***})$	&	$(-0.67,1.1E-06^{***})$	&	$(-1.05,1.8E-11^{***})$	\\
Adj Type(appearance)	&	$(-0.32,2.7E-02^*)$	&	$(0.16,2.8E-01)$	&	$(-0.10,5.3E-01)$	\\
You(M)	&	$(2.12,2.0E-13^{***})$	&	$(1.62,2.0E-07^{***})$	&	$(0.95,8.2E-03^{**})$	\\
You(F)	&	$(-4.06,4.4E-14^{***})$	&	$(-3.27,4.3E-28^{***})$	&	$(-6.02,3.3E-09^{***})$	\\
Lookahead(M)	&	$(1.83,4.4E-13^{***})$	&	$(0.26,3.7E-01)$	&	$(1.47,1.8E-04^{***})$	\\
Lookahead(F)	&	$(-2.48,3.2E-21^{***})$	&	$(-2.68,1.3E-22^{***})$	&	$(-1.69,2.8E-11^{***})$	\\
Consistency(M)	&	$(-0.10,6.7E-01)$	&	$(-2.23,1.1E-15^{***})$	&	$(0.18,6.2E-01)$	\\
Consistency(F)	&	$(0.25,3.0E-01)$	&	$(0.64,1.9E-03^{**})$	&	$(0.67,3.7E-03^{**})$	\\
Opposite(M)	&	$(3.56,7.1E-59^{***})$	&	$(4.31,6.0E-50^{***})$	&	$(3.28,2.1E-40^{***})$	\\
\hline
\end{tabular}
    \caption{Structural factors regression analysis for the systems with the greatest opposite binary gender tendency (\textbf{Spanish}). As above (refer to Table \ref{tab:adverb_reg_es}), the variable to predict in the adjective declension choice, where female is the positive class. Unlike the prior regression results, here we include determined-gender cases in order to assess the difficulty introduced by different structural factors. Therefore, the true gender of the referent must be controlled for (\textbf{True(M)} is 1 when the true label is $M$, 0 when the true label is $F$). In addition to the regular adjective traits, we include structural factors consistency(M/F): 1 if an earlier adjective refers to the same entity and is M/F, lookahead(M/F): 1 if an adjective's referent appears for the first time after the adjective and the known gender is M/F, and you(M/F): 1 if the adjective refers to ``you'' and the known gender is M/F. Lookahead and you variables must be paired with the true label because they affect the task difficulty regardless of gender. The results show that both lookahead and you strongly increase difficulty (as indicated by strong, significant, and positive coefficients when the correct label is $M$, and conversely strong, significant, and negative coefficients when the correct label is $F$). That is, the coefficients indicate an increased likelihood of choosing the incorrect gender agreement, while all else is controlled for. The variable of interest in these systems is the \textbf{Opposite(M)}: 1 when the other referent in conversation is known to be male. Systems with a strong opposite binary gender effect have strong positive coefficients, indicating an increased likelihood of a $F$-adjective.}
    \label{tab:tab:struct_reg_es}
\end{table*}

\begin{table*}
\centering
\begin{tabular}{l l l l}
\hline
\textbf{Variable} & \textbf{ONLINE-A} & \textbf{CUNI-GA} & \textbf{TranssionMT} \\
\hline
Intercept	&	$(4.82,4.9E-30^{***})$	&	$(18.93,9.5E-01)$	&	$(4.57,4.5E-35^{***})$	\\
True(M)	&	$(-12.42,3.3E-145^{***})$	&	$(-27.04,9.2E-01)$	&	$(-10.64,1.3E-147^{***})$	\\
Adj Stereo(M)	&	$(-0.01,9.6E-01)$	&	$(-0.30,1.2E-01)$	&	$(-0.24,1.7E-01)$	\\
Adj Stereo(F)	&	$(0.99,8.6E-15^{***})$	&	$(0.89,1.3E-12^{***})$	&	$(0.83,2.8E-13^{***})$	\\
Adj Sentiment(neg)	&	$(-0.49,1.5E-05^{***})$	&	$(-0.61,2.0E-08^{***})$	&	$(-0.44,1.1E-05^{***})$	\\
Adj Type(appearance)	&	$(0.16,2.3E-01)$	&	$(0.58,3.2E-06^{***})$	&	$(0.45,1.3E-04^{***})$	\\
You(M)	&	$(2.61,1.6E-39^{***})$	&	$(4.55,6.8E-32^{***})$	&	$(2.47,1.4E-33^{***})$	\\
You(F)	&	$(-6.18,4.4E-44^{***})$	&	$(-20.47,9.4E-01)$	&	$(-5.99,1.9E-52^{***})$	\\
Lookahead(M)	&	$(2.25,2.8E-22^{***})$	&	$(1.06,1.7E-06^{***})$	&	$(1.50,9.8E-13^{***})$	\\
Lookahead(F)	&	$(-1.38,2.2E-07^{***})$	&	$(-0.96,5.8E-04^{***})$	&	$(-0.72,1.7E-03^{**})$	\\
Consistency(M)	&	$(0.33,6.1E-02)$	&	$(0.46,3.1E-02^*)$	&	$(0.25,1.8E-01)$	\\
Consistency(F)	&	$(0.22,2.7E-01)$	&	$(0.44,9.4E-02)$	&	$(0.09,6.6E-01)$	\\
Opposite(M)	&	$(4.80,1.9E-127^{***})$	&	$(3.13,3.8E-98^{***})$	&	$(3.10,1.7E-116^{***})$	\\
\hline
\end{tabular}
    \caption{Structural factors regression analysis for the systems with the greatest opposite binary gender tendency (\textbf{Czech}).}
    \label{tab:tab:struct_reg_cs}
\end{table*}

\begin{table*}
\centering
\begin{tabular}{l l l l}
\hline
\textbf{Variable} & \textbf{AMI} & \textbf{ONLINE-A} & \textbf{TranssionMT} \\
\hline
Intercept	&	$(3.89,1.1E-128^{***})$	&	$(1.34,1.3E-34^{***})$	&	$(3.17,2.5E-89^{***})$	\\
True(M)	&	$(-22.31,9.4E-01)$	&	$(-4.71,3.8E-127^{***})$	&	$(-6.73,3.9E-174^{***})$	\\
Adj Stereo(M)	&	$(-0.44,1.4E-03^{**})$	&	$(-0.01,9.2E-01)$	&	$(-0.94,6.0E-09^{***})$	\\
Adj Stereo(F)	&	$(0.34,7.1E-04^{***})$	&	$(0.78,1.3E-15^{***})$	&	$(0.39,1.2E-04^{***})$	\\
Adj Sentiment(neg)	&	$(-0.18,3.9E-02^*)$	&	$(0.09,3.0E-01)$	&	$(0.15,9.4E-02)$	\\
Adj Type(appearance)	&	$(-0.09,3.7E-01)$	&	$(0.75,1.6E-12^{***})$	&	$(0.04,7.2E-01)$	\\
You(M)	&	$(19.86,9.4E-01)$	&	$(2.19,3.5E-21^{***})$	&	$(3.15,5.2E-38^{***})$	\\
You(F)	&	$(-3.78,3.0E-79^{***})$	&	$(-3.21,4.0E-81^{***})$	&	$(-4.68,3.0E-90^{***})$	\\
Lookahead(M)	&	$(-1.26,1.0E-10^{***})$	&	$(0.15,4.2E-01)$	&	$(-0.38,5.3E-02)$	\\
Lookahead(F)	&	$(0.16,4.1E-01)$	&	$(0.85,1.7E-05^{***})$	&	$(0.73,1.8E-03^{**})$	\\
Consistency(M)	&	$(-0.79,8.1E-07^{***})$	&	$(-0.73,1.4E-09^{***})$	&	$(-0.05,7.5E-01)$	\\
Consistency(F)	&	$(-0.40,3.4E-02^*)$	&	$(-0.48,3.9E-03^{**})$	&	$(0.57,9.4E-03^{**})$	\\
Opposite(M)	&	$(-1.78,2.0E-75^{***})$	&	$(-0.96,1.4E-26^{***})$	&	$(-0.99,2.4E-24^{***})$	\\
\hline
\end{tabular}
    \caption{Structural factors regression analysis for the systems with the greatest opposite binary gender tendency (\textbf{Icelandic}).}
    \label{tab:struct_reg_is}
\end{table*}

\end{document}